%% file: root.tex
\documentclass[letterpaper, 10 pt, conference]{ieeeconf}  

\IEEEoverridecommandlockouts                              

\usepackage[font=small]{caption}
\usepackage{graphicx} 
\usepackage{tabularx}
\usepackage{times} 
\usepackage{amsmath} 
\usepackage{amssymb}  
\usepackage{comment}
\usepackage{url}
\usepackage{subfigure}
\usepackage{multirow}
\usepackage{colortbl}
\usepackage{booktabs}
\usepackage[table,xcdraw]{xcolor}
\usepackage[normalem]{ulem}
\newcolumntype{x}[1]{>{\centering\arraybackslash}p{#1pt}}
\newcolumntype{y}[1]{>{\raggedright\arraybackslash}p{#1pt}}
\newcolumntype{z}[1]{>{\raggedleft\arraybackslash}p{#1pt}}
\usepackage{xcolor}
\usepackage{xspace}

\makeatletter
\let\NAT@parse\undefined
\makeatother
\usepackage[pagebackref=false,breaklinks=true,letterpaper=true,colorlinks,bookmarks=false]{hyperref}
\newcommand{\tablestyle}[2]{\setlength{\tabcolsep}{#1}\renewcommand{\arraystretch}{#2}\centering\footnotesize}
\useunder{\uline}{\ul}{}

\makeatletter
\DeclareRobustCommand\onedot{\futurelet\@let@token\@onedot}
\def\@onedot{\ifx\@let@token.\else.\null\fi\xspace}

\definecolor{socialnav_blue}{RGB}{67, 109, 183}
\definecolor{socialnav_red}{RGB}{231, 66, 52}
\definecolor{socialnav_yellow}{RGB}{251, 189, 5}
\definecolor{socialnav_green}{RGB}{51, 168, 82}
\definecolor{socialnav_gray}{RGB}{165, 165, 165}
\definecolor{lblue}{RGB}{66, 133, 244}

\title{\LARGE \bf
 NavThinker: Action-Conditioned World Models for Coupled  \\Prediction and Planning in Social Navigation
}

\author{Tianshuai Hu$^{1}$, Zeying Gong$^{2}$, Lingdong Kong$^{3}$, XiaoDong Mei$^{1}$, Yiyi Ding$^{2}$, \\ Qi Zeng$^{2}$, Ao Liang$^{4}$, Rong Li$^{2}$, Yangyi Zhong$^{2}$ and Junwei Liang$^{1,2*}$%
\thanks{$^{1}$ The Hong Kong University of Science and Technology. {\tt\footnotesize \{thuaj, xmeiab\}@connect.ust.hk}}%
\thanks{$^{2}$ The Hong Kong University of Science and Technology (Guangzhou). 
{\tt\footnotesize \{zgong313, ydingaz, qzeng721, rli335, yzhong123\}@connect.hkust-gz.edu.cn, junweiliang@hkust-gz.edu.cn}}%
\thanks{$^{3}$ National University of Singapore. {\tt\footnotesize lingdong.kong@u.nus.edu}}%
\thanks{$^{4}$ University of Chinese Academy of Sciences {\tt\footnotesize liangao@sia.cn}}%
\thanks{* Corresponding author.}
}

\begin{document}

\maketitle
\thispagestyle{empty}
\pagestyle{empty}

\input{sections/0_abstract}
\input{sections/1_intro}
\input{figs/fig2}
\input{sections/2_related_work}
\input{sections/3_formulation}
\input{sections/4_methodology}
\input{sections/5_experiments}

\input{sections/6_conclusion}

\bibliographystyle{IEEEtran}
\bibliography{IEEEabrv,ref}

\end{document}

%% file: sections/0_abstract.tex
\begin{abstract}
    Social navigation requires robots to act safely in dynamic human environments. Effective behavior demands \emph{thinking ahead}: reasoning about how the scene and pedestrians evolve under different robot actions rather than reacting to current observations alone. This creates a coupled prediction-planning challenge, where robot actions and human motion mutually influence each other. To address this challenge, we propose \textbf{NavThinker}, a future-aware framework that couples an action-conditioned world model with on-policy reinforcement learning. The world model operates in the Depth Anything V2 patch feature space and performs autoregressive prediction of future scene geometry and human motion; multi-head decoders then produce future depth maps and human trajectories, yielding a future-aware state aligned with traversability and interaction risk. Crucially, we train the policy with DD-PPO while injecting world-model think-ahead signals via: (i) action-conditioned future features fused into the current observation embedding and (ii) social reward shaping from predicted human trajectories. Experiments on single- and multi-robot Social-HM3D show state-of-the-art navigation success, with zero-shot transfer to Social-MP3D and real-world deployment on a Unitree Go2, validating generalization and practical applicability. Webpage:~\url{https://hutslib.github.io/NavThinker/}.
\end{abstract}

%% file: sections/1_intro.tex
\section{Introduction}
\label{sec:intro}

Social navigation requires robots to reach goals safely while respecting human comfort and social norms in dynamic crowds~\cite{mavrogiannis2023core,francis2025principles,kabir2026socially}. In such environments, robot and human motions are mutually coupled~\cite{mavrogiannis2023core,samavi2024sicnav,sun2021move}. Effective navigation therefore demands \emph{thinking ahead}: reasoning about how the scene will evolve under different actions rather than reacting to observations alone.

\input{figs/fig1}

Most existing approaches lack ``look-ahead'' ability. Prior methods either choose actions from instantaneous observations, behaving myopically in tight interactions~\cite{helbing1995social,ferrer2013robot,fiorini1998motion,van2008reciprocal,mavrogiannis2023core}; treat pedestrian forecasts as fixed inputs independent of robot actions and plan against these predictions~\cite{thomas2023foreseeable,cui2021learning,eiffert2020path,poddar2023crowd}; or jointly reason over robot and human futures via expensive optimization that scales poorly with many agents and long horizons~\cite{samavi2024sicnav,sun2021move}. A promising solution lies in world models within a reinforcement learning (RL) framework. Unlike standard RL, which learns interaction-aware policies through implicit reasoning from experience, a world model explicitly generates future latent features conditioned on actions, thereby modeling interactive dynamics.

This approach, known as RL-based world modeling \cite{hafner2023mastering,shanks2025dreamernav,oguzie2023predictive} offers a natural mechanism for action-conditioned foresight: by learning environment dynamics, it enables imagination that couples prediction and planning within a single learned model. However, bringing world models to social navigation exposes two key challenges: \emph{task-misaligned spatial representations} and \emph{limited mechanisms to inject foresight into the policy}.

The first challenge is that many latent world models do not capture the task-relevant factors needed for safe social navigation. Success depends on both \emph{scene geometry} and \emph{human motion} to represent traversability and interaction risk. We address this by operating the world model in the Depth Anything V2 (DA-V2) patch feature space~\cite{yang2024depth}, and using depth and trajectory decoders to anchor representations to geometry and motion, improving imagined depth fidelity and enabling trajectory-informed policy learning.

The more fundamental challenge, however, is how to effectively inject this foresight into the policy for future-aware planning. Social navigation requires anticipating human responses to each candidate action and planning accordingly, yet current approaches fall short. Pure imagination methods~\cite{hafner2023mastering,hafner2020mastering} accumulate compounding errors in complex scenes~\cite{janner2019trust}. Other approaches, whether using imagination for representation learning~\cite{liu2025x} or reward augmentation~\cite{hirose2024selfi}, keep prediction and planning largely decoupled, that is, they do not query the world model for lookahead planning. We achieve this coupling by two \textit{think-ahead mechanisms} that form the core of our approach: fusing action-conditioned future features into the observation embedding, and shaping rewards from predicted human trajectories. Experiments show that each mechanism progressively improves navigation success.

Overall, NavThinker achieves state-of-the-art performance on Social-HM3D, surpassing the previous future-aware method~\cite{gong2025cognition} by 3.2\% in success rate and 2.95\% in SPL, while reducing human collisions. We further extend the benchmark to a more complex multi-robot setting where multiple robots independently use the policy to reach their goals, and NavThinker consistently outperforms all baselines. NavThinker also generalizes zero-shot to Social-MP3D and transfers to a real Unitree Go2 robot. 

To summarize, the primary contributions of this work are:
\begin{itemize}
    \item We propose \textbf{NavThinker}, a future-aware framework that couples an action-conditioned world model with policy optimization. The world model performs autoregressive prediction in the Depth Anything V2 (DA-V2) patch feature space; depth and trajectory decoders then align learned representations with traversability and interaction risk.
    
    \item We introduce two think-ahead mechanisms that inject world-model predictions into policy optimization: \emph{Action-Conditioned Future-Feature Fusion} as lookahead context, and \emph{Social Reward Shaping} from predicted human trajectories.
    
    \item We demonstrate state-of-the-art results on single- and multi-robot Social-HM3D, zero-shot generalization to Social-MP3D, and transfer to a real Unitree Go2.
\end{itemize}

%% file: figs/fig1.tex
\begin{figure}[t]
\centering
\includegraphics[width=0.95\linewidth]{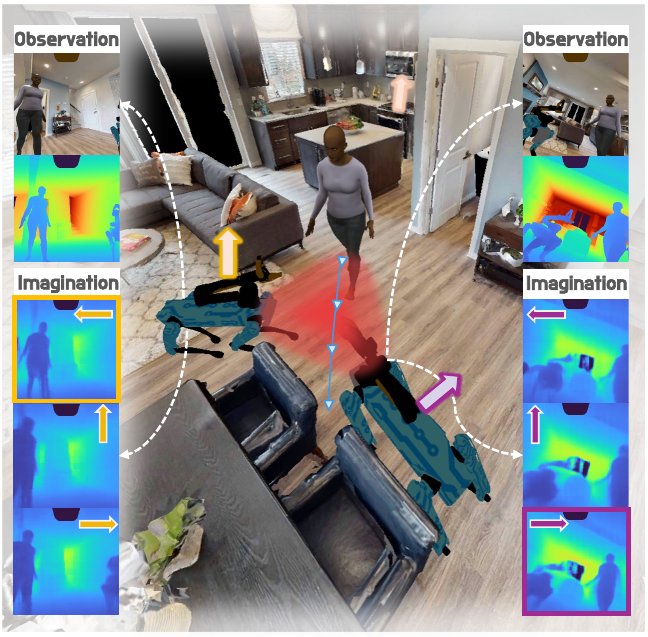}
\caption{\textbf{NavThinker: future-aware social navigation.} Social navigation requires the robot to reach a goal in environments shared with dynamic pedestrians. Red regions highlight interaction zones where the robot and pedestrians may conflict. Top: egocentric depth observations. Bottom: the world model imagines action-conditioned futures for the corresponding observations. By imagining how the scene evolves under different candidate actions, NavThinker anticipates potential social conflicts and selects safe, socially compliant actions.}
\label{fig:teaser}
\vspace{-2em}
\end{figure}

%% file: figs/fig2.tex
\begin{figure*}[t]
\centering
\includegraphics[width=0.83\linewidth]{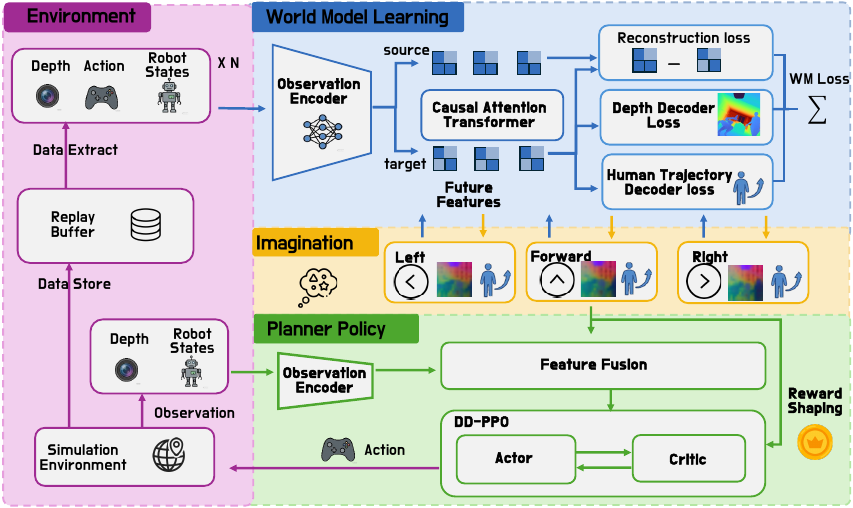}
\caption{\textbf{Overview of the NavThinker framework.} Our framework consists of two modules: a world model that learns action-conditioned scene dynamics (top), and an imagination-augmented planner policy trained with DD-PPO (bottom). During \textbf{World Model Learning}, depth observations, actions, and robot states are extracted from the Replay Buffer. A frozen DA-V2 ViT encoder extracts patch embeddings, and a Causal Attention Transformer autoregressively predicts future latent features. A Depth Decoder and a Human Trajectory Decoder are trained alongside a latent consistency loss to anchor representations to scene geometry and human motion. During \textbf{Policy Learning}, the Observation Encoder produces the current embedding from depth and robot states. The Imagination module queries the world model under each candidate action, generating action-conditioned future features that are fused with the current embedding via Feature Fusion. The fused representation feeds into the DD-PPO Actor-Critic network. Predicted human trajectories additionally provide Reward Shaping, coupling prediction with planning in an imagine-then-act loop.}
\label{fig:workfolw}
\vspace{-1.5em}
\end{figure*}

%% file: sections/2_related_work.tex
\section{Related Work}
\label{related_works}

\subsection{Social Navigation \& Prediction-Planning Integration}
In crowded social environments, navigation is governed by interaction dynamics rather than independent agent motions. The bidirectional influence between robot actions and pedestrian responses makes prediction and planning inherently coupled~\cite{mavrogiannis2023core,francis2025principles,kabir2026socially}. Reactive approaches select actions solely from current observations, including social-force models and reciprocal collision avoidance~\cite{helbing1995social,van2010optimal}, as well as local trajectory optimization~\cite{rosmann2013efficient}. Although efficient, they are myopic and exhibit freezing or oscillatory behavior in dense crowds~\cite{mavrogiannis2023core}. Proactive methods incorporate future reasoning by predicting evolving free space or pedestrian trajectories~\cite{thomas2023foreseeable,poddar2023crowd}. However, many remain decoupled: predictors are trained independently and their outputs are treated as fixed inputs to downstream planners, limiting feedback from planning to prediction. More tightly coupled formulations integrate forecasting and control through bilevel optimization or model-based RL~\cite{samavi2024sicnav}. While narrowing the prediction--planning gap, optimization-based coupling is often computationally heavy, and latent world-model approaches may compress spatial information into global vectors, obscuring fine-grained geometry and interaction structure. Our work targets this coupled regime while preserving spatial fidelity and enabling stable on-policy training.

\subsection{World Models for Embodied Decision-Making}
World models learn predictive environment dynamics to support planning or policy optimization~\cite{ha2018world,hafner2019learning}. Dreamer-style recurrent state-space models compress observations into global latent vectors and train policies via imagination~\cite{hafner2023mastering}, achieving strong sample efficiency but discarding explicit spatial structure. Recent work improves temporal modeling with Transformers~\cite{zhang2023storm} or preserves spatial detail through diffusion-based prediction~\cite{alonso2024diffusion}. Diffusion models, however, require multi-step denoising at inference, limiting rollout efficiency. Other approaches predict features from frozen vision backbones such as DINOv2~\cite{oquab2023dinov2} to enable efficient planning~\cite{zhou2024dino}, yet often lack explicit temporal modeling or interaction-aware objectives. In contrast, we build an action-conditioned world model in the DA-V2 patch feature space, preserving spatial structure via patch tokens while modeling temporal dependencies with a block-causal Transformer. Joint prediction of depth, human trajectories, and rewards from a shared latent state unifies geometry and interaction dynamics within a single predictive framework.

\subsection{Imagination-Based Reinforcement Learning}
World models shape policy learning in several ways. Pure imagination actor--critic methods train policies entirely from latent rollouts~\cite{hafner2020mastering,hafner2023mastering}, improving sample efficiency but increasing exposure to compounding model bias. Representation-augmentation approaches inject learned latent features into model-free policies while keeping optimization grounded in real interaction~\cite{liu2025x}, yet do not exploit action-conditioned multi-step imagination. Reward or data augmentation modifies training signals using imagined rollouts or synthetic transitions~\cite{hirose2024selfi}, improving efficiency without fully coupling prediction and planning. Our approach retains image-based, on-policy PPO training~\cite{schulman2017proximal} while enhancing it through two mechanisms: detached late fusion of predictive features, and action-conditioned imagination that forecasts human responses and injects socially aware penalties into the reward. Unlike latent-only training, the policy remains grounded in real observations; unlike static auxiliary prediction, imagined futures depend explicitly on candidate actions, enabling closed-loop coupling between prediction and planning without sacrificing stability.

%% file: sections/3_formulation.tex
\section{Problem Formulation}
\label{sec:problem_formulation}
We consider a point-goal SocialNav task in environments populated by $N$ dynamic humans. Due to egocentric sensing, occlusions, and limited field-of-view, the environment state can only be partially observed. We formulate this task as a \emph{Partially Observable Markov Decision Process} (POMDP)
$\left(\mathcal{S}, \mathcal{O}, \mathcal{A}, \mathcal{T}, r, \gamma\right)$,
where $s_t\in\mathcal{S}$ is the latent environment state, $o_t\in\mathcal{O}$ is the observation, $a_t\in\mathcal{A}$ is the action, $\mathcal{T}(s_{t+1}\!\mid s_t,a_t)$ is the transition kernel, $r(s_t,a_t)$ is the reward, and $\gamma\in(0,1)$ is the discount factor.
At each timestep $t$, the agent selects $a_t$ based on the observation history $o_{\le t}$.

\subsection{State and Observation}
We decompose the latent state, for each timestamp $t$, into the robot state, static scene, and human states:
\begin{equation}
s_t \triangleq \big(x_t,\; \mathcal{M},\; \{h_t^{(i)}\}_{i=1}^{N}\big),
\end{equation}
where $x_t$ denotes the robot pose, $\mathcal{M}$ denotes the static scene structure, and $h_t^{(i)}$ denotes the $i$-th human's pose.
The robot receives a partial observation as:
\begin{equation}
o_t \triangleq \big(d_t,\; g,\; x_t),
\end{equation}
where $d_t$ is the egocentric depth image, $g$ specifies the navigation goal, and $x_t$ the robot's own pose.
With limited observability, neither the full scene $\mathcal{M}$ nor the complete human states $\{h_t^{(i)}\}_{i=1}^{N}$ are directly observed.

\subsection{Action Space}
The robot executes a discrete action set $a_t\in\mathcal{A}$, where
$\mathcal{A}=\{\textsc{Forward},\,\textsc{Turn-Left},\,\textsc{Turn-Right},\,\textsc{Stop}\}$.

\subsection{Objective and the Coupling Challenge}
We seek a policy $\pi(a_t\mid o_{\le t})$ that maximizes the expected return:
\begin{equation}
\begin{aligned}
\pi^\star
=\arg\max_{\pi}\;
\mathbb{E}\Big[\sum\nolimits_{t=0}^{\infty}\gamma^t\, r(s_t,a_t)\Big],& \\
\text{with}\quad
s_{t+1}\sim \mathcal{T}(\cdot\mid s_t,a_t),\;
a_t\sim \pi(\cdot\mid o_{\le t}).&
\end{aligned}
\label{eq:pomdp_objective}
\end{equation}
Equivalently, from a planning view, the agent selects actions to trade off task efficiency and social compliance under evolving human motion:
\begin{equation}
\begin{aligned}
\mathbf{a}^\star
=\arg\min_{\mathbf{a}\in\mathcal{A}^H}
\Big(c_{\mathrm{task}}(\mathbf{a})
+\lambda\, c_{\mathrm{social}}(\mathbf{a}, \boldsymbol{\xi}_{1:N})\Big).
\end{aligned}
\label{eq:socnav_planning}
\end{equation}
Here $\mathbf{a}$ denotes a length-$H$ action sequence, $\boldsymbol{\xi}_{1:N}$ denotes future human trajectories, $c_{\mathrm{task}}$ captures progress-to-goal and path efficiency, and $c_{\mathrm{social}}$ penalizes social violations.

This exposes the key difficulty in social navigation: effective decisions require reasoning about \emph{future human motion}, coupling prediction and planning under partial observability.

\subsection{Latent World Modeling for SocialNav}
\label{sec:latent_wm}
NavThinker couples a \emph{world model} and a \emph{policy} in a closed loop over a latent feature state $Z_t$.
The state is inferred from observations, which is:
\begin{equation}
\label{eq:belief_update_latent}
Z_t=f_{\mathrm{enc}}(o_t).
\end{equation}
Given $Z_t$, the world model imagines the next latent state under each candidate action $a\in\mathcal{A}$:
\begin{equation}
\label{eq:wm_latent}
\hat{Z}_{t+1}^{(a)}=\mathcal{W}(Z_t,\,a).
\end{equation}
The policy then selects an action conditioned on both the current observation and the imagined futures:
\begin{equation}
\label{eq:policy_latent}
a_t=\pi_\theta\!\bigl(o_t,\;\{\hat{Z}_{t+1}^{(a)}\}_{a\in\mathcal{A}},\;g\bigr).
\end{equation}
This imagine-then-act cycle enables the robot to anticipate how the scene will evolve under each possible action, coupling prediction and planning under partial observability.

%% file: sections/4_methodology.tex
\section{Methodology}
\label{sec:method}
\subsection{Overview}
\label{sec:overview}
NavThinker tackles coupled prediction and planning in social navigation by learning a world model and using it as an imagination engine for policy improvement.
As shown in Fig.~\ref{fig:workfolw}, it consists of two modules:
(i)~an action-conditioned \textbf{scene-and-interaction world model} operating in DA-V2 patch-feature space, which predicts future patch latents and uses multi-head decoders to forecast future depth and human trajectories, producing a future-aware state capturing traversability and interaction risk;
and (ii)~an \textbf{imagination-augmented planner policy} trained on-policy with DD-PPO, which exploits imagined rollouts by fusing candidate-action-conditioned future features with the current observation embedding and by shaping rewards from predicted human trajectories.
This imagination-enabled training couples prediction and planning through the learned latent dynamics.

\subsection{Scene-and-Interaction World Model Learning}
\label{sec:wm}
Social navigation is partially observable, with incomplete perception of scene structure and human states; moreover, robot actions and human motion are bidirectionally coupled.
We therefore learn a latent world model in the patch-feature space with a frozen Depth Anything V2 (DA-V2) encoder~\cite{yang2024depth}, predicting the action-conditioned evolution of the scene.
Our model consists of three components:
\begin{align}
\text{Observation model:}\quad & z_t = \mathrm{enc}(o_t), \label{eq:obs_model}\\
\text{Transition model:}\quad & \hat{z}_{t+1} \sim p_\phi(\hat{z}_{t+1} \mid z_{t-H:t},\, a_{t-H:t}), \label{eq:trans_model}\\
\text{Decoder heads:}\quad & \hat{d}_t,\;\hat{\boldsymbol{\xi}}_{1:N_h},\;\hat{r}_t = q_\psi(z_t), \label{eq:decoder_model}
\end{align}
where $z_t\in\mathbb{R}^{P\times D}$ denotes the patch embeddings ($P$ patches, $D$ dimensions), $H$ is the context length, and the decoder heads produce depth~ $\hat{d}_t$, human trajectory~$\hat{\boldsymbol{\xi}}_{1:N_h}$, and reward~$\hat{r}_t$ predictions respectively.

\paragraph{Observation encoder}
We adopt a frozen DA-V2 ViT~\cite{yang2024depth} as the observation model.
At each timestep $t$, it encodes the depth image $o_t$ into patch embeddings $z_t\in\mathbb{R}^{P\times D}$.
The encoder remains frozen during training, providing task-independent, spatially rich representations that generalise across environments.

\paragraph{Transition model}
We adopt a ViT-based causal Transformer~\cite{dosovitskiy2020image} operating directly on patch embeddings.
Each action $a_t$ is encoded into a $D$-dimensional token and appended to the patch sequence, yielding $(P{+}1)$ tokens per frame.
A causal sliding-window mask of length $H$ restricts each frame to attend only to its preceding context, enabling frame-level autoregressive prediction that captures global scene structure and temporal coherence.
The transition model is trained with a latent consistency loss:
\begin{equation}
\label{eq:latent_consistency}
\mathcal{L}_{f}(\phi) = \|\hat{z}_{t+1} - z_{t+1}\|_2^2,
\end{equation}
which aligns the predicted patch features $\hat{z}_{t+1}$ with the frozen encoder's output $z_{t+1}$.

\paragraph{Decoder heads}
Task-specific decoders predict depth, human trajectories, and reward from the predicted patch features; their training is independent of the transition model.
The \emph{depth decoder} follows a DPT-style architecture~\cite{ranftl2021vision} that reassembles patch tokens into multi-scale feature maps and progressively upsamples them to reconstruct the normalised depth map:
\begin{equation}
\label{eq:depth_dec}
\hat{d}_t = \mathrm{DPT}(z_t) \in [0,1]^{H_d \times W_d}.
\end{equation}
The \emph{trajectory decoder} takes the global feature $\mathrm{MeanPool}(z_t)$ and predicts human's future positions over a horizon~$T$:
\begin{equation}
\label{eq:traj_dec}
\hat{\boldsymbol{\xi}}_{1:N_h} = \mathrm{TrajDec}\!\bigl(\mathrm{MeanPool}(z_t)\bigr).
\end{equation}
The \emph{reward decoder} predicts a scalar reward from the same global feature: $\hat{r}_t = \mathrm{MLP}\!\bigl(\mathrm{MeanPool}(z_t)\bigr)$.
The combined decoder loss is:
\begin{equation}
\label{eq:decoder_loss}
\mathcal{L}_{\mathrm{dec}}(\psi) = \mathcal{L}_{d}(d_t, \hat{d}_t) + \lambda_\xi\,\mathcal{L}_{\xi}(\boldsymbol{\xi}_{1:N_h}, \hat{\boldsymbol{\xi}}_{1:N_h}) + \lambda_r\,\mathcal{L}_{r}(r_t, \hat{r}_t).
\end{equation}

\paragraph{Training objective}
The transition model is trained with a latent consistency loss between predicted and ground-truth patch features.
The decoders are trained independently with task-specific losses.
Depth reconstruction anchors the belief to local geometry, human forecasting captures interaction dynamics, and the latent consistency term ensures faithful imagination:
\begin{equation}
\label{eq:wm_loss}
\mathcal{L}_{\mathrm{WM}}(\phi)
=\underbrace{\lambda_f\,\|\hat{z}_{t+1}-z_{t+1}\|_2^2}_{\text{latent consistency}}
+\underbrace{\mathcal{L}_{d}+\lambda_\xi\,\mathcal{L}_{\xi}+\lambda_r\,\mathcal{L}_{r}}_{\text{decoder losses}},
\end{equation}
where $\mathcal{L}_d$, $\mathcal{L}_\xi$, $\mathcal{L}_r$ supervise depth reconstruction, human trajectory forecasting, and reward prediction, respectively.
The latent consistency term aligns imagined patch features with the frozen encoder's output.

\input{tables/single-robot-benchmark}
\input{tables/multi-robot-benchmark}

\subsection{Imagination-Augmented Policy Learning}
\label{sec:policy}
A learned world model that captures scene geometry and human dynamics enables the policy to \emph{think before acting}: by imagining the consequences of each candidate action in latent space, the agent can anticipate future interactions without additional environment steps.
We train an imagination-augmented policy that queries the world model to anticipate action consequences before execution.
Our policy consists of three components:
\begin{align}
\text{Observation Encoder:}\quad & e_t = \mathrm{Enc}(o_t,\, a_{t-1}), \label{eq:obs_enc}\\
\text{Imagination Module:}\quad & l_t = \mathrm{Imagine}(z_{t-H:t},\, \mathcal{A};\, \phi), \label{eq:imagine}\\
\text{Policy Head:}\quad & a_t,\, V_t = \pi_\theta(e_t,\, l_t,\, g),\; V_\psi(e_t,\, l_t,\, g), \label{eq:policy_head}
\end{align}
where $e_t \in \mathbb{R}^{d_e}$ denotes the observation embedding, $l_t \in \mathbb{R}^{|\mathcal{A}| \cdot d_l}$ aggregates imagined outcomes for all candidate actions, and $g$ is the goal specification.

\paragraph{Observation encoder}
A ResNet encoder~\cite{he2016deep} extracts visual features from the depth image $o_t$, which are concatenated with the previous action $a_{t-1}$ and fused through a GRU to produce the observation embedding $e_t \in \mathbb{R}^{d_e}$. This recurrent encoding captures temporal context beyond the current frame.

\paragraph{Imagination module}
We query the world model's transition model $p_\phi$ to predict latent outcomes for each candidate action:
\begin{equation}
\label{eq:lookahead}
\hat{z}_{t+1}^{(a)} = p_\phi(z_{t-H:t},\, a), \quad \forall\, a \in \mathcal{A}.
\end{equation}
The predicted latent states are concatenated to form the lookahead feature $l_t$ and then fusion with the observation embedding $e_t$, enabling the policy to anticipate future interactions before acting.

\paragraph{Policy head}
The policy head comprises an actor that outputs a distribution over discrete actions and a critic that estimates state values:
\begin{align}
a_t &\sim \pi_\theta(\cdot \mid e_t,\, l_t,\, g), \label{eq:actor}\\
V_\psi &= V_\psi(e_t,\, l_t,\, g). \label{eq:critic}
\end{align}

\paragraph{Reward design and training objective}
We train the policy with DD-PPO~\cite{wijmans2019dd} under a reward that combines task progress with world-model-derived social shaping:
\begin{equation}
\label{eq:reward}
R_t = \underbrace{r_t^{\mathrm{goal}} + r_t^{\mathrm{succ}}}_{\text{task progress}}
    - \underbrace{r_t^{\mathrm{coll}}}_{\text{immediate safety}}
    - \underbrace{r_t^{\mathrm{traj}}}_{\text{predictive social cost}},
\end{equation}
where $r_t^{\mathrm{goal}}$ measures geodesic progress toward the goal, $r_t^{\mathrm{succ}}$ provides a terminal bonus upon success, and $r_t^{\mathrm{coll}}$ penalizes collisions with static obstacles and humans. The predictive social cost $r_t^{\mathrm{traj}}$ is computed from the action-conditioned forecasts of human trajectories $\hat{\boldsymbol{\xi}}_{1:N_h}$, coupling human motion prediction with decision-making: the world model predicts future human positions under candidate actions, and the policy is optimized to maintain safe separation from these predicted interaction futures. This encourage anticipatory and socially compliant behaviors.

%% file: tables/single-robot-benchmark.tex
\begin{table*}[!ht]
\centering
\caption{\textbf{Single-Robot Social Navigation Results} on Social-HM3D and Social-MP3D (zero-shot transfer from Social-HM3D). We report SR/SPL (success/efficiency), PSC (social compliance), and H-Coll (human collisions). Best and 2nd-best are in \textbf{bold} and \underline{underline}.}
\vspace{-0.1cm}
\label{tab:single_benchmark}
\tablestyle{7pt}{1.1}
\begin{tabular}{
>{\raggedright\arraybackslash}p{2.4cm}
>{\centering\arraybackslash}p{1.3cm}
>{\centering\arraybackslash}p{1.3cm}
>{\centering\arraybackslash}p{1.3cm}
>{\centering\arraybackslash}p{1.3cm}
|
>{\centering\arraybackslash}p{1.3cm}
>{\centering\arraybackslash}p{1.3cm}
>{\centering\arraybackslash}p{1.3cm}
>{\centering\arraybackslash}p{1.3cm}
}
\toprule
\multirow{2}{*}{\textbf{Methods}} 
& \multicolumn{4}{c}{\textbf{Social-HM3D}} 
& \multicolumn{4}{c}{\textbf{Social-MP3D}} 
\\
\cmidrule(lr){2-5}
\cmidrule(lr){6-9}

& SR$\uparrow$ & SPL$\uparrow$ & PSC$\uparrow$ & H-Coll$\downarrow$
& SR$\uparrow$ & SPL$\uparrow$ & PSC$\uparrow$ & H-Coll$\downarrow$
\\
\midrule

\rowcolor{gray!15}
\multicolumn{9}{l}{\textbf{Rule-based}}
\\
A*~\cite{hart1968formal} & 44.81 & 43.99 & \underline{90.38} & 54.80  & 45.67  & \underline{44.69}  & 91.97  & 54.00  \\
ORCA~\cite{van2011reciprocal} & 37.44 & 32.91 & \textbf{92.23} & \underline{39.77} & 38.81 & 34.65 & \textbf{94.03} & \underline{39.86} \\

\midrule

\rowcolor{gray!15}
\multicolumn{9}{l}{\textbf{Reinforcement Learning-based}}
\\
Habitat-official~\cite{puig2023habitat} & 38.99 & 33.53 & 90.37 & 55.48 & 37.00 & 31.76 & 92.03 & 52.33 \\
Falcon~\cite{gong2025cognition}          & \underline{56.26} & \underline{52.05} & 89.76 & 41.22  & \textbf{51.67} & \textbf{45.54} & 92.53 & 40.67 \\

\midrule

\rowcolor{blue!15}
\textbf{NavThinker}  & \textbf{59.46} & \textbf{55.00} & 89.91 & \textbf{39.09} & \underline{47.33}  & 41.71 & \underline{93.68} & \textbf{37.67} 
\\
\bottomrule
\end{tabular}
\vspace{-1.5em}
\end{table*}

%% file: tables/multi-robot-benchmark.tex
\begin{table}[t]
\centering
\caption{\textbf{Multi-Robot Social Navigation Results} on Social-HM3D. Robots navigate to individual goals without communication. We report SR/SPL (success/efficiency), PSC (social compliance), H-Coll (human collisions), and team-level T-SR/T-SPL. Best and second-best are highlighted in \textbf{bold} and \underline{underline}, respectively.}
\vspace{-0.1cm}
\label{tab:multi_benchmark}
\tablestyle{2pt}{1}
\resizebox{\linewidth}{!}{
\begin{tabular}{
>{\centering\arraybackslash}p{2.4cm}
>{\centering\arraybackslash}p{0.9cm}
>{\centering\arraybackslash}p{0.9cm}
>{\centering\arraybackslash}p{0.9cm}
>{\centering\arraybackslash}p{1.0cm}
>{\centering\arraybackslash}p{0.9cm}
>{\centering\arraybackslash}p{0.9cm}
}
\toprule
Method 
& SR$\uparrow$ 
& SPL$\uparrow$ 
& PSC$\uparrow$ 
& H-Coll$\downarrow$ 
& T-SR$\uparrow$ 
& T-SPL$\uparrow$ 
\\
\midrule
\rowcolor{gray!15}
\multicolumn{7}{l}{\textbf{Rule-Based}} \\

A*~\cite{hart1968formal}  
& 26.06 & 25.70 & 95.20 & 35.68 & 14.76 & 14.51 \\

ORCA~\cite{van2011reciprocal} 
& 24.13 & 22.48 & 95.53 & 35.13 & 15.09 & 13.84 \\

\midrule

\rowcolor{gray!15}
\multicolumn{7}{l}{\textbf{Reinforcement Learning}} \\

Habitat-official~\cite{puig2023habitat} 
& 26.78 & 24.41 & 95.47 & 31.39 & 14.98 & 13.68 \\

Falcon~\cite{gong2025cognition} 
& \underline{28.63} & \underline{26.48} & \underline{94.98} & \underline{28.63} & \underline{16.08} & \underline{15.12} \\

\midrule

\rowcolor{blue!15}
\textbf{NavThinker}
& \textbf{30.04}
& \textbf{28.14}
& \textbf{95.55}
& \textbf{25.55}
& \textbf{16.30}
& \textbf{15.22} 
\\
\bottomrule
\end{tabular}}
\vspace{-1em}
\end{table}

%% file: sections/5_experiments.tex
\section{Experiments}
\label{sec:exps}
In this section, we first outline evaluation details and then present quantitative comparisons with state-of-the-art methods on single-robot and multi-robot benchmarks. A series of ablation studies are presented to evaluate the individual components of our approach: the two think-ahead mechanisms and our world model quality. We then report zero-shot generalization and real-world deployment results, and conclude with qualitative analysis.

\begin{figure*}[t]
    \centering
    \includegraphics[width=0.85\linewidth]{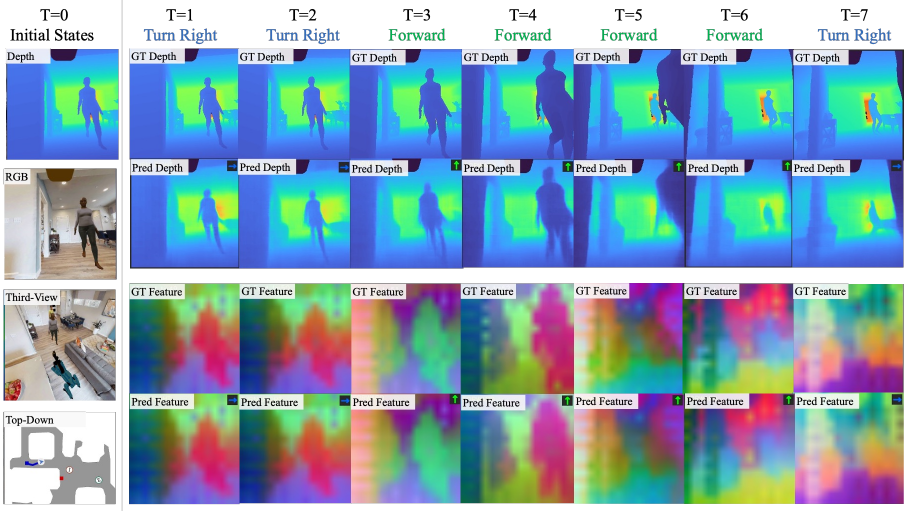}
    \vspace{-0.1cm}
    \caption{\textbf{World model imagination quality.} Left: current depth observation with RGB, third-person view, and top-down map for reference. Right: after executing the action at the previous timestep, we show (from top to bottom) the ground-truth depth, the world model's predicted depth, the PCA visualization of ground-truth DA-V2 features, and the PCA visualization of predicted features. The world model produces faithful depth predictions and latent features that closely match the ground truth across different actions.}
    \label{fig:world_model_reconstruction_quality_vis}
    \vspace{-1em}
\end{figure*}

\subsection{Experiments Setup}

\noindent\textbf{Benchmarks.}
We evaluate NavThinker on the Social-HM3D benchmark~\cite{gong2025cognition}, built upon photo-realistic Habitat environments populated with goal-driven, collision-aware pedestrians. Each episode places controllable robots alongside autonomous humans; every robot must navigate to its goal while respecting personal space and avoiding collisions. We consider two settings: (i)~\textbf{single-robot}, where one robot navigates among dynamic humans, and (ii)~\textbf{multi-robot}, a more challenging setting we introduce to increase benchmark complexity, where multiple robots independently execute their own policies to reach individual goals without communication. To evaluate generalization, we additionally test on \textbf{Social-MP3D} in a zero-shot transfer setting.

\noindent\textbf{Metrics.}
Following~\cite{gong2025cognition}, we report per-robot metrics: Success Rate~(SR), Success Weighted by Path-Length~(SPL), Personal Space Compliance~(PSC), and Human Collision rate~(H-Coll). For the multi-robot setting, we additionally report Team Success Rate~(T-SR), which equals 1 only when \emph{all} robots reach their goals, and Team SPL~(T-SPL), which rewards path efficiency only under full team success.

\noindent\textbf{Baselines.}
We evaluate NavThinker against rule-based planners A*~\cite{hart1968formal} and ORCA~\cite{van2011reciprocal}, RL methods from Habitat-official~\cite{puig2023habitat}, and the future-aware baseline Falcon~\cite{gong2025cognition}.

\subsection{Quantitative Results}

\paragraph{Single-robot social navigation.}
As shown in Tab.~\ref{tab:single_benchmark}, NavThinker achieves state-of-the-art performance with the highest SR (59.46) and SPL (55.00) while maintaining comparable personal space compliance. Compared to the future-aware method Falcon, NavThinker improves success rate by 3.2\% and SPL by 2.95\%, while reducing human collisions from 41.22\% to 39.09\%, demonstrating that thinking ahead enables both efficient and socially compliant navigation.

\input{tables/policy_ablation}
\input{tables/policy_ablation_multi}
\input{tables/world-model-reconstruction}

\paragraph{Multi-robot social navigation.}
To further evaluate NavThinker in more complex coordination scenarios, we extend Social-HM3D with a multi-robot setting where multiple robots independently use their own policies to reach individual goals. As shown in Tab.~\ref{tab:multi_benchmark}, NavThinker consistently outperforms all baselines across both per-robot and team-level metrics, demonstrating that NavThinker is capable of handling more complex scenarios with denser interactions.

\paragraph{Zero-shot generalization.}
To assess generalization ability, we directly evaluate models trained on Social-HM3D on Social-MP3D without any fine-tuning. Tab.~\ref{tab:single_benchmark} reports results testing on 300 episodes of Social-MP3D. NavThinker maintains competitive performance in this zero-shot transfer setting, demonstrating that the learned world model and think-ahead mechanisms generalize across different environments and scene layouts.

\input{figs/fig3}
\begin{figure}[h]
\centering
\includegraphics[width=0.95\linewidth]{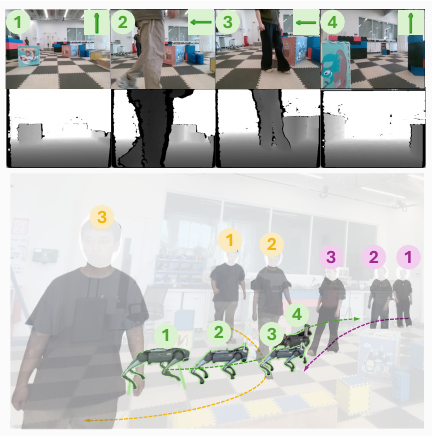}
\vspace{-0.2cm}
\caption{\textbf{Real-world deployment on a Unitree Go2 robot.} In this scenario, the robot navigates around a group of people and reaches its goal while maintaining safe distances.}
\label{fig:real_world}
\vspace{-1.5em}
\end{figure}

\subsection{Ablation Study \& Analyses}
We conduct ablation studies to validate the two key challenges and corresponding contributions: (i)~the think-ahead mechanisms for injecting foresight into the policy, and (ii)~the task-aligned world model with depth and trajectory decoders.
\subsubsection{Policy: Think-Ahead Mechanisms}
\label{sec:ablation_policy}

To validate that injecting foresight into the policy enables coupled prediction and planning, we ablate the two think-ahead mechanisms: lookahead imagination (LookH.) and trajectory-based social reward shaping (TrajR.).

\paragraph{Single-robot setting.}
As shown in Tab.~\ref{tab:policy_ablation_single}, the base policy without any think-ahead mechanism achieves only 43.36\% SR. Adding lookahead imagination raises SR to 55.87\% (+12.51\% relative), demonstrating that fusing action-conditioned future features into the observation embedding substantially improves navigation by enabling the policy to compare imagined outcomes before acting. Further incorporating trajectory-based reward shaping yields 59.46\% SR (+3.59\% relative over lookahead alone), confirming that shaping rewards from predicted human trajectories encourages the policy to anticipate social interactions.

\paragraph{Multi-robot setting.}
The same trend is observed in the multi-robot setting (Tab.~\ref{tab:policy_ablation_multi}), where each mechanism again progressively improves both success rate and collision avoidance. The improvements are even more pronounced in this setting, confirming that the think-ahead mechanisms are especially important in complex scenarios where anticipating the behavior of both humans and other robots is critical.

\subsubsection{World Model: Task-Aligned Decoders}
\label{sec:ablation_wm}

To validate that task-aligned decoders improve the quality of imagined futures, we evaluate world model variants with different decoder configurations on a held-out replay buffer. As shown in Tab.~\ref{tab:wm_ablation}, adding the depth decoder $\mathcal{L}_d$ improves cosine similarity between predicted and target latent features from 0.9278 to 0.9342, and reduces depth RMSE from 0.0406 to 0.0389. Further incorporating the trajectory decoder $\mathcal{L}_\xi$ yields the best latent consistency (CosSim 0.9383) and depth fidelity (RMSE 0.0371), confirming that anchoring the world model to both scene geometry and human motion produces higher-quality imagined futures. 

Notably, the trajectory decoder's contribution extends beyond world model fidelity: the trajectory-based reward shaping directly consumes predicted human trajectories from the world model, and its consistent improvement on policy performance (Tab.~\ref{tab:policy_ablation_single},~\ref{tab:policy_ablation_multi}) provides evidence that the task-aligned decoder benefits not only imagination quality but also downstream decision-making.

\subsection{Qualitative Analysis}
Figure~\ref{fig:qualitative} presents qualitative comparisons between NavThinker and baseline methods in representative social navigation scenarios on Social-HM3D.

\subsection{Real-World Deployments}
To validate the practical applicability of our approach in real environments, we deploy NavThinker on a Unitree Go2 quadruped robot. We use an Intel RealSense D455i depth camera as the depth input to the model. The system operates in a client-server architecture: the Go2 handles sensor data acquisition and locomotion control on-board, while streaming depth observations to a server machine for model inference. The predicted actions are sent back to the Go2 for execution. NavThinker successfully navigates to designated goals while maintaining safe distances from pedestrians.

%% file: tables/policy_ablation.tex
\begin{table}[t]
\centering
\caption{\textbf{Ablation Study} of NavThinker planning policy components on the single-robot Social-HM3D benchmark. LookH.: world-model lookahead imagination; TrajR.: trajectory-based social reward shaping. Metrics include SR/SPL (success/efficiency), PSC (social compliance), and H-Coll (human collisions).}
\vspace{-0.1cm}
\label{tab:policy_ablation_single}
\tablestyle{2pt}{1.1}
\begin{tabular}{
>{\centering\arraybackslash}p{0.9cm}
>{\centering\arraybackslash}p{0.9cm}|
>{\centering\arraybackslash}p{1.2cm}
>{\centering\arraybackslash}p{1.2cm}
>{\centering\arraybackslash}p{1.2cm}
>{\centering\arraybackslash}p{1.2cm}
}
\toprule
LookH. & TrajR. 
& SR$\uparrow$ & SPL$\uparrow$ & PSC$\uparrow$ & H-Coll$\downarrow$
\\
\midrule

$\times$ & $\times$
& 43.36 & 36.27 & 90.81 & 47.33
\\

$\checkmark$ & $\times$
& 55.87 & 49.36 & 90.58 & 39.19 
\\

\rowcolor{blue!15}
$\checkmark$ & $\checkmark$
& 59.46 & 55.00 & 89.91 & 39.09
\\
\bottomrule
\end{tabular}
\vspace{-0.5em}
\end{table}

%% file: tables/policy_ablation_multi.tex
\begin{table}[t]
\centering
\caption{\textbf{Ablation Study} of planner policy components on multi-robot Social-HM3D. LookH.: world-model lookahead imagination; TrajR.: trajectory-based social reward shaping. Metrics include SR/SPL (success/efficiency), PSC (social compliance), H-Coll (human collisions), and T-SR/T-SPL (team-level success/efficiency).}
\vspace{-0.1cm}
\label{tab:policy_ablation_multi}
\tablestyle{2pt}{1}
\begin{tabular}{
>{\centering\arraybackslash}p{0.7cm}
>{\centering\arraybackslash}p{0.7cm}|
>{\centering\arraybackslash}p{1cm}
>{\centering\arraybackslash}p{1cm}
>{\centering\arraybackslash}p{1cm}
>{\centering\arraybackslash}p{1cm}
>{\centering\arraybackslash}p{1cm}
>{\centering\arraybackslash}p{1cm}
}
\toprule
LookH. & TrajR.
& SR$\uparrow$
& SPL$\uparrow$
& PSC$\uparrow$
& H-Coll$\downarrow$
& T-SR$\uparrow$
& T-SPL$\uparrow$

\\
\midrule

$\times$ & $\times$
& 13.64 & 11.27 & 93.85 & 59.25 & 13.44 & 11.13 \\

$\checkmark$ & $\times$
& 28.52 & 26.52 & 95.47 & 25.88 & 14.76 & 13.73 \\

\rowcolor{blue!15}
$\checkmark$ & $\checkmark$
& 30.04 & 28.14 & 95.55 & 25.55 & 16.30 & 15.22 
\\
\bottomrule
\end{tabular}
\vspace{-0.5em}
\end{table}

%% file: tables/world-model-reconstruction.tex
\begin{table}[t]
\label{tab:wm_ablation}
\centering
\caption{\textbf{Ablation Study} on auxiliary losses for World Model training. Evaluation on 200 episodes from the held-out replay buffer. $\mathcal{L}_d$ is the depth reconstruction loss and $\mathcal{L}_\xi$ is the human trajectory prediction loss. CosSim is the cosine similarity between predicted and target latent features. Depth RMSE is the root mean squared error of predicted depth images. Traj ADE/FDE are the average and final displacement errors of predicted human trajectories.}
\vspace{-0.1cm}
\label{tab:wm_ablation}
\tablestyle{2pt}{1.1}
\begin{tabular}{
>{\centering\arraybackslash}p{0.6cm}
>{\centering\arraybackslash}p{0.6cm}|
>{\centering\arraybackslash}p{1.3cm}
>{\centering\arraybackslash}p{1.8cm}
>{\centering\arraybackslash}p{1.5cm}
>{\centering\arraybackslash}p{1.5cm}
}
\toprule
$\mathcal{L}_{d}$ & $\mathcal{L}_\xi$
& CosSim$\uparrow$ & Depth RMSE$\downarrow$ & Traj ADE$\downarrow$ & Traj FDE$\downarrow$ \\
\midrule

$\times$ & $\times$
& 0.9278 & 0.0406 & --- & --- \\
$\checkmark$ & $\times$
& 0.9342 & 0.0389 & --- & --- \\
$\checkmark$ & $\checkmark$
& 0.9383 & 0.0371 & 1.74 & 1.93 \\

\bottomrule
\end{tabular}
\vspace{-2em}
\end{table}

%% file: figs/fig3.tex
\begin{figure}[h]
\centering
\includegraphics[width=0.85\linewidth]{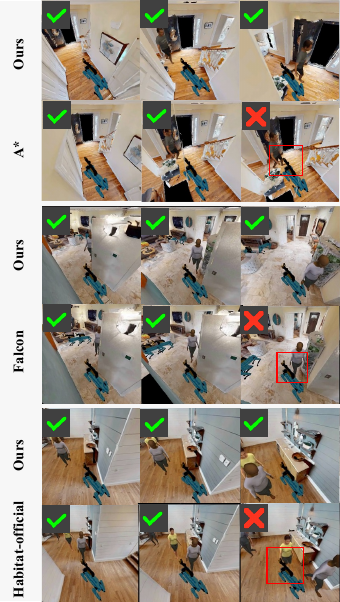}
\caption{\textbf{Qualitative comparisons on Social-HM3D.} Navthinker produces safe actions under different social situations.
\textbf{Rows 1–2 (Blind corner):} A* collides when turning, while ours avoids the pedestrian.
\textbf{Rows 3–4 (Intersection):} Falcon collides at the intersection, while ours safely yields and passes.
\textbf{Rows 5–6 (Front approach):} Habitat-official collides in a head-on encounter, while ours gives way for the pedestrian.
Red boxes mark baseline failures; green checkmarks indicate safe behaviors.}
\label{fig:qualitative}
\vspace{-1.5em}
\end{figure}

%% file: sections/6_conclusion.tex
\section{Conclusion}
\label{sec:conclusion}
In this paper, we proposed NavThinker, a future-aware social navigation framework that couples a task-aligned world model in the DA-V2 patch feature space with on-policy reinforcement learning. Depth and trajectory decoders anchor representations to scene geometry and human motion, while two think-ahead mechanisms, namely future feature fusion and social reward shaping, inject foresight into the policy for coupled prediction and planning. NavThinker achieves state-of-the-art results on single-robot and multi-robot Social-HM3D, generalizes zero-shot to Social-MP3D, and transfers to a real Unitree Go2. Future research may concentrate on extending the world model to incorporate multi-modal observations (\emph{e.g.}, RGB and semantic information) for richer scene understanding, or exploring longer imagination horizons to handle more complex social interactions.